\documentclass[conference]{IEEEtran}
\IEEEoverridecommandlockouts

\usepackage{cite}
\usepackage{amsmath,amssymb,amsfonts}
\usepackage{graphicx}
\usepackage{textcomp}
\usepackage{xcolor}
\def\BibTeX{{\rm B\kern-.05em{\sc i\kern-.025em b}\kern-.08em
    T\kern-.1667em\lower.7ex\hbox{E}\kern-.125emX}}

\usepackage{graphicx}
\usepackage{graphics}
\usepackage{subfigure} 
\usepackage{multirow}
\usepackage{float}
\usepackage{booktabs}
\usepackage{amsthm}
\theoremstyle{definition}

\usepackage{amssymb}
\usepackage{calligra}
\usepackage{algorithm}
\usepackage{algorithmicx}
\usepackage{algpseudocode} 
\usepackage{amsmath}

\begin{document}

\title{SemDP: Semantic-level Differential Privacy Protection for Face Datasets
}

\author{\IEEEauthorblockN{Xiaoting Zhang}
\IEEEauthorblockA{\textit{Nanjing University of Aeronautics and Astronautics }\\
Nanjing, China \\
xiaotingz1992@nuaa.edu.cn}
\and
\IEEEauthorblockN{Tao Wang}
\IEEEauthorblockA{\textit{Nanjing University of Aeronautics and Astronautics}\\
Nanjing, China \\
wangtao21@nuaa.edu.cn}
\and
\IEEEauthorblockN{Junhao Ji}
\IEEEauthorblockA{\textit{Nanjing University of Aeronautics and Astronautics}\\
Nanjing, China \\
jijunhao@nuaa.edu.cn}

}

\maketitle

\begin{abstract}
While large-scale face datasets have advanced deep learning-based face analysis, they also raise privacy concerns due to the sensitive personal information they contain. Recent schemes have implemented differential privacy to protect face datasets. However, these schemes generally treat each image as a separate database, which does not fully meet the core requirements of differential privacy. In this paper, we propose a semantic-level differential privacy protection scheme that applies to the entire face dataset. Unlike pixel-level differential privacy approaches, our scheme guarantees that semantic privacy in faces is not compromised. The key idea is to convert unstructured data into structured data to enable the application of differential privacy. Specifically, we first extract semantic information from the face dataset to build an attribute database, then apply differential perturbations to obscure this attribute data, and finally use an image synthesis model to generate a protected face dataset. Extensive experimental results show that our scheme can maintain visual naturalness and balance the privacy-utility trade-off compared to the mainstream schemes.
\end{abstract}


\section{Introduction}
Huge amounts of face data are ubiquitously captured every day from a variety of recording devices such as smartphones and personal cameras. A face image dataset contains a large number of faces along with their corresponding label information, which serves as a substantial resource within the domain of computer vision. Owing to the widespread availability of face image data, deep learning has demonstrated remarkable advancements across multiple domains within the realm of artificial intelligence. The continual enhancement in the scale and quality of face image datasets has led to new challenges and opportunities for vision tasks based on data-driven optimization, such as face recognition, expression analysis, and age-gender identification.

Through the analysis of pixel information, texture, contours, and other inherent characteristics within images, sensitive personal attributes can be extracted, such as gender, age, ethnicity, and facial expressions. This can cause privacy leakages and lead to misuse. Moreover, some datasets may contain geographic location information of the objects or their biometric characteristics, which can further exacerbate privacy risks. In light of these circumstances, privacy preservation in face data publishing emerges as an essential research topic, aiming to enable learning from data while protecting personally sensitive information.

The field of face protection has attracted considerable attention in the current landscape of information technology. Homomorphic encryption \cite{ref1, ref2} is an encryption technique in which the original data is encrypted and then subjected to subsequent computational operations on the encrypted data. However, this process demands substantial computational resources. Secure multiparty computation (SMC) \cite{ref3, ref4} enables multiple parties to engage in computational tasks on encrypted data, ensuring that no party can access sensitive information held by other participants. A notable drawback of SMC is the computational overhead associated with encryption protocols, making it less computationally efficient compared to other methods. Anonymization \cite{ref5, ref6} encompasses the process of concealing or modifying personally identifiable information with the aim of preventing the re-identification of individuals in a dataset. Nevertheless, malicious adversaries could employ advanced techniques to reverse the anonymization process and reveal individual identities. Confusion techniques \cite{ref7, ref8, ref9, ref10} provide a layer of privacy protection in data-driven environments by introducing noise or distortion into datasets. However, excessive confusion may result in data utility loss, thus affecting the accuracy and reliability of analytical results. Moreover, a common weakness of all the aforementioned methods lies in the complexity of quantifying their effectiveness in protecting the original data. This complexity prevents the establishment of a theoretical foundation to guarantee the level of privacy protection.

To overcome this difficulty, one might resort to differential privacy (DP) \cite{ref11, ref12}, which provides a mechanism for guaranteeing and quantifying system privacy through a robust mathematical formulation. Differential privacy has become the predominant norm for statistically analyzing databases containing sensitive data. The robustness of differential privacy is grounded in its meticulous mathematical formulation, which ensures privacy without compromising the model's accuracy for reliable statistical inference. Differential privacy is a concept proposed for database privacy, which addresses the statistical distinguishability of data and ensures the privacy of individual data. That is, differential privacy offers a methodology for extracting valuable insights from sensitive data while mitigating the risk of individual-specific information disclosure. Therefore, even if an attacker has arbitrary background knowledge, the privacy risk can be significantly reduced.

DP-Pix \cite{ref13, ref14} first integrates the differential privacy framework into the realm of image processing, introducing the concept of adjacent databases for neighboring images. DP-Pix effectively prevents the reconstruction of original pixels. However, the current works cannot guarantee semantic privacy; that is, it is possible to reconstruct specific semantics within the image, which introduces a risk of privacy leakage. In the realm of DP-Pix, the protection primarily focuses on the pixels within the image. For face images, it is important to highlight that privacy protection does not focus on the pixels themselves; rather, it centers on preserving semantics. In addition, the results in existing works resemble conventional pixelization or blurring techniques, leading to lower naturalness and utility compared to desirable standards. This is due to the considerable amount of differential privacy noise necessary to effectively obscure the original image.

Instead of treating image pixels as database records, DP-Semantic \cite{ref16, ref17, ref18, ref19} provides DP guarantees for the representation of images using semantic latent codes and subsequently generates a novel image by utilizing the privatized latent space representation. The randomized manipulations applied within the latent semantic space are capable of preserving essential attributes inherent to the face image. However, these existing works do not meet the definition of differential privacy. For example, the work proposed in \cite{ref19} aims to protect the identification of people in face images. There exists a balance between the preservation of privacy and the level of utility in the context of differential privacy. In other words, it is crucial to maintain accessibility while preserving identity. This implies that the work in \cite{ref19} may lead to a certain degree of identity leakage.

\begin{figure*}[]
	\centering
	\vspace{1.0em}
	\includegraphics[width=0.9\linewidth]{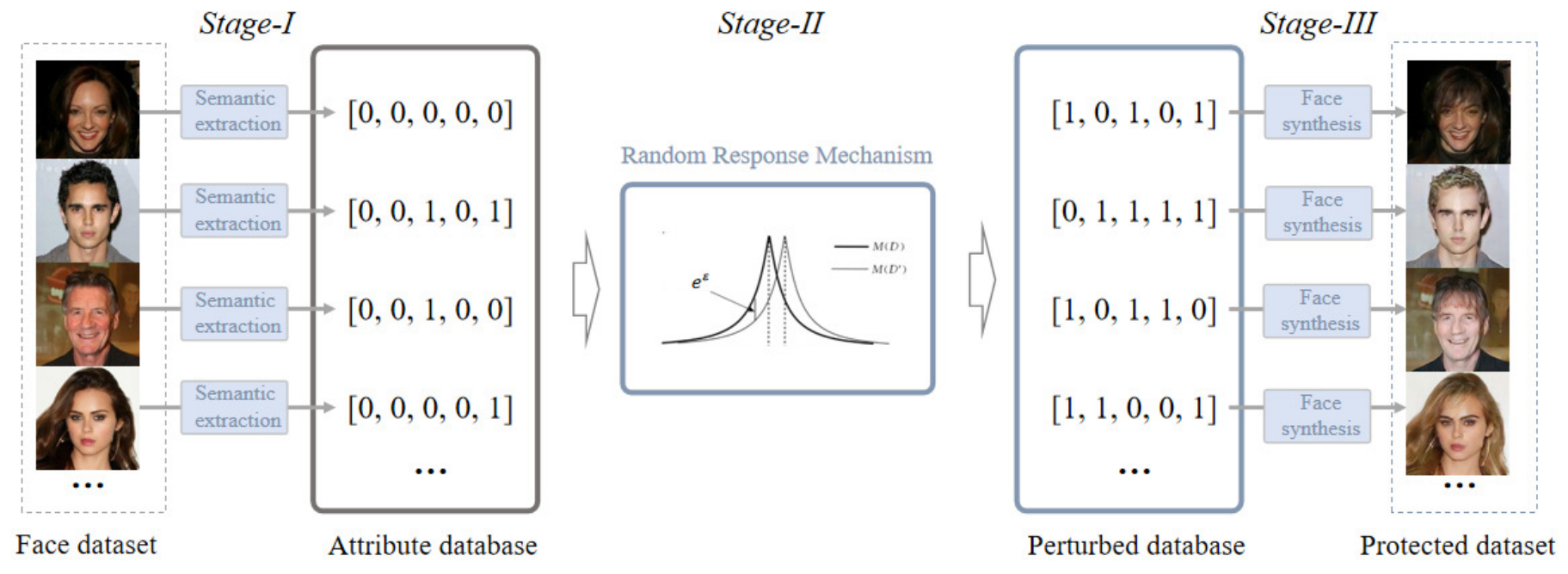}
	\caption{The workflow of the proposed scheme. Stage-I: Constructing the face attribute database corresponding to the face image dataset. Stage-II: Generating the released face attribute database under the randomized response mechanism. Stage-III: Implementing the image synthesis technique to obtain the released face image dataset.}
	\label{fig1}
\end{figure*}

Nevertheless, the previous works mainly suffer from a limitation in considering an image as the database. This fails to satisfy the definition of differential privacy and cannot guarantee both semantic privacy and utility. To overcome this limitation, we propose a semantic-level differential privacy protection scheme for the face image dataset. In the context of the face image dataset, our objective is to publish aggregated insights from the face attribute database, while concurrently minimizing the risk of exploiting it to infer sensitive semantic information about individual records. Our scheme accomplishes this by generating a perturbed attribute database using the randomized response mechanism. The proposed scheme comprises three stages, as illustrated in the workflow shown in Fig.\ref{fig1}. Stage-I aims to construct the face attribute database for the face image dataset. Stage-II implements a randomized response mechanism, incorporating adjustable perturbations based on differential privacy into the face attribute database. Stage-III performs a face image dataset synthesis technique, which employs a well-trained generative adversarial network (GAN) to generate face images using the perturbed face attribute database.

The principal contributions are outlined as follows: 
\begin{itemize}
	\item We design a semantic-level differential privacy protection scheme, which can provide provable protection for the face dataset. 
	\item We consider the whole face dataset (not a single face) as a database, which satisfies the basic requirements of differential privacy.  
	\item Our scheme ensures that the semantics in the face are not disclosed, which cannot be guaranteed by pixel-level differential privacy.
\end{itemize}

The following sections of this paper are structured as follows. We present a concise summary of related work in Section II. The preliminaries are formalized in Section III, including relevant theories of differential privacy and randomized response. Three stages of our scheme are elaborated in Section IV. Sufficient experimental results are shown in Section V, while Section VI concludes with an examination of potential directions for future research.

\section{Related Work}
\subsection{Face dataset protection}
The extensive use of multimedia technologies like video surveillance, online meetings, and drones has made it easier to gather a large amount of face image data, thereby raising significant concerns about privacy. Recent methods for protecting face privacy can be classified into two main categories: face identity privacy protection and face attribute privacy protection.

Face identity privacy protection entails the preservation of facial attributes while simultaneously eliminating identity information from the input images. Early studies \cite{ref20, ref21} primarily utilize methods such as masking, pixelization, and blurring to obfuscate face parts. These techniques suffer from the overprotection problem, which results in unnecessary information loss. In recent times, innovative techniques have been devised to enhance the privacy of facial information. Sun \textit{et al.} \cite{ref22} attempted to conceal the facial region in the image and subsequently generate a new face through inpainting. CIAGAN \cite{ref23} ensures that the generated image exhibits distinct identities with comparable features with the source image, relying on landmark information and obscured images. Furthermore, some researchers have focused on the advancement of reversible anonymization methods. Gu et al. \cite{ref24} trained a conditional GAN with multi-task learning objectives, where the input image and password are used as conditions to generate the corresponding anonymized image. Cao \textit{et al.} \cite{ref34} proposed the integration of adaptable parameters and user-specific passwords to accomplish invertible de-identification. RiDDLE \cite{ref35} effectively encrypts and decrypts facial identity in the latent embedding of the pretrained StyleGAN2 architecture.

The objective of face attribute privacy protection is to preserve the utility of identities while simultaneously concealing privacy attributes from third-party applications. Mirjalili \textit{et al.} \cite{ref36} devised a semi-adversarial training network designed to perturb face images with the goal of obscuring gender attributes. Gender-AN \cite{ref37} utilizes an encoder-decoder network based on GAN to perturb face images and is independent of specific attributes. Xie \textit{et al.} \cite{ref38} developed a reversible scheme by employing iteratively refined adversarial perturbations to prevent gender identification. These approaches primarily rely on adversarial perturbations, in which the protected results are notably affected by the target model and may exhibit artifacts. Several representation-level methods have been proposed to obtain a face representation that excludes privacy-sensitive attributes. Morales \textit{et al.} \cite{ref39} introduced a privacy-preserving feature representation, which aims to protect the privacy of specific attributes. RAPP \cite{ref40} exhibits the capability to conceal a range of attributes within images and subsequently restore original images.

\subsection{Image differential privacy}
Image data, especially when it contains facial elements, has the potential to unveil personal and sensitive information. The consideration of privacy issues becomes imperative when sharing image data with untrusted third parties. While differential privacy has been acknowledged as the predominant principle in data privacy protection, it becomes a challenge to directly apply standard DP to non-aggregate data. Our focus is on summarizing the applications of differential privacy within the domain of image data privacy.

DP-Pix \cite{ref13, ref14} represents a pioneering advancement in the extension of DP to the realm of individual-level image publication. This approach introduces the concept of neighboring images for content protection and reduces sensitivity via pixelization. However, this approach pixelizes the entire image, resulting in low image quality. To enhance the utility of obfuscated images, Fan \cite{ref15} further proposed DP-SVD, which transforms the given image into a feature vector and then perturbs the vector with noise that satisfies metric-DP before performing the reverse transformation. Based on DP-SVD, Yan \textit{et al.} \cite{ref41} proposed CODER, which adopts an improved distortion metric definition that measures the distance more precisely to improve utility. This distortion metric can be applied to an arbitrary k-dimensional metric space with stronger image privacy protection. Reilly \textit{et al.} \cite{ref31} adapted a video sanitization method to individual images, which consist of pixel clustering, allocation of budget, pixel sampling, and interpolation techniques. Chamikara \textit{et al.} \cite{ref25} implemented differential privacy perturbations on eigenfaces, with a uniform allocation of the privacy budget to each individual eigenface. This allocation strategy contributed to a significant reduction in accuracy. Saleem \textit{et al.} \cite{ref27} introduced DP-Shield, an interactive framework designed for the obfuscation of face images, operating under the stringent principles of differential privacy.

To maintain a level of image semantics, Li and Clifton \cite{ref16} suggested incorporating differential privacy guarantees into the representation of image latent vectors. Within this approach, privacy allocations were distributed among various elements in the latent space and sensitivity was clipped into the maximum observed bounds. Liu \textit{et al.} \cite{ref17} applied DP to image feature vectors extracted by an encoder network and then utilizes a GAN to produce obfuscated images based on these perturbed features. This approach lacks a formal guarantee of privacy due to its reliance on empirical sensitivity estimates. IdentityDP \cite{ref19} is a framework aimed at anonymizing face data, which uses a differential privacy mechanism in the deep neural network. Cao \textit{et al.} \cite{ref42} anonymized faces by averaging the attributes of the k nearest neighbors and then applies perturbations with differential privacy. Their results exhibit numerous artifacts and lack realism. Considering perceptual similarity, Chen \textit{et al.} \cite{ref18} introduced perceptual indistinguishability (PI) as a standardized notion of image privacy, and then proposed PI-Net to accomplish image obfuscation. Based on differential privacy, Ji \textit{et al.} \cite{ref26} designed a learnable mechanism for allocating privacy budgets, and put forward a framework for protecting face privacy.

\section{Preliminary Work}
\subsection{Differential privacy}
Differential privacy (DP) quantifies the degree of protection for individual privacy, providing a robust mathematical framework for evaluating and ensuring the strength of privacy protection. The initial formalization of this concept is designed to offer statistical guarantees, ensuring that the published data cannot disclose the presence or absence of individuals within the dataset. In other words, an algorithm is deemed differentially private if and only if, for any pair of input datasets that vary by the incorporation or exclusion of an individual participant, the probability distributions of the algorithm's outputs exhibit remarkable similarity. This indicates that even when adversaries possess knowledge of all the data except for one individual, their capacity to deduce specific information regarding that particular individual remains constrained. These datasets are defined as neighboring datasets.

\textbf{Definition 1 ($\varepsilon $-Differential Privacy).} A randomized mechanism $\mathcal{M}$ adheres to $\varepsilon $-DP, if for any neighbor-pair datasets ${D}$ and ${D}'$, and all $S\subseteq Range\left( \mathcal{M} \right)$, 
\begin{equation}
	\Pr \left[ \mathcal{M}\left( D \right)\in S \right]\le {{e}^{\varepsilon }}\Pr \left[ \mathcal{M}\left( {{D}'} \right)\in S \right]. \label{1}
\end{equation}
If ${S}$ is a countable set, then we can modify Eq. (1) as 
\begin{equation}
	\Pr \left[ \mathcal{M}\left( D \right)=s \right]\le {{e}^{\varepsilon }}\Pr \left[ \mathcal{M}\left( {{D}'} \right)=s \right].  \label{2}
\end{equation}
Here, we consider ${S}$ as a singleton set.

The privacy budget, denoted as $\varepsilon $, signifies an upper limit on the ratio of probabilities associated with the occurrence of identical outputs in neighboring datasets. As the $\varepsilon $ value decreases, the strength of the privacy assurance increases. To release useful responses, differential privacy functions incorporate controlled noise into the database using a randomization mechanism, achieving the required degree of similarity among potential configurations of the database contents.

\subsection{Randomised response}
The randomized response mechanism is introduced to address the following survey challenge: to ascertain the proportion of individuals within the population who exhibit a specific sensitive attribute $\mathcal{A}$. By convention, the attribute is binary; a value of 1 signifies the presence of the sensitive attribute, whereas 0 indicates its absence in the respondent. The randomized response mechanism involves two distinct participant groups, the respondents $\mathcal{R}=\left\{ {{R}_{1}},{{R}_{2}},\cdots ,{{R}_{n}} \right\}$ and the questioner ${Q}$. Each respondent ${{R}_{i}}$ has an actual answer ${{x}_{i}}\in {0,1}$ with respect to the binary attribute $\mathcal{A}$. In the context of a survey regarding privacy sent by the questioner ${Q}$, the respondents are unwilling to disclose their individual information ${{x}_{i}}$. To protect privacy, each respondent ${{R}_{i}}$ conceals his or her personal sensitive information by giving a randomized response. The response ${{y}_{i}}\in {0,1}$ of an individual respondent is considered a randomized representation of his or her true answer ${{x}_{i}}\in {0,1}$.

Consistent with standard notation, $\left( \Omega ,\mathcal{F},Pr \right)$ denotes a probability space. For each $i\in [n]$, the random variable ${{y}_{i}}$ mapped from the sample space $\Omega$ to the set ${0,1}$ is conditional upon the accurate value ${{x}_{i}}$. The randomized response mechanism is defined by
\begin{equation}
	\Pr \left( {{y}_{i}}=v|{{x}_{i}}=u \right)={{p}_{uv}},  \label{3}
\end{equation}
which leads to the formulation of the design matrix. Here, ${{p}_{uv}}\in (0,1)$, $u$ denotes the random output, and $v$ signifies the true attribute value.

\textbf{Definition 2 (Design Matrix)}. The randomized response mechanism on a sensitive binary attribute, as defined in Eq. (3), is distinctly characterized by its $2\times 2$ design matrix,
\begin{equation}
	P=\left( \begin{matrix}
		{{p}_{00}} & {{p}_{01}}  \\
		{{p}_{10}} & {{p}_{11}}  \\
	\end{matrix} \right),  \label{4}
\end{equation}
where ${{p}_{00}},{{p}_{01}},{{p}_{10}},{{p}_{11}}\in (0,1)$. To ensure that the probability mass functions of each ${{y}_{i}}$ sum up to 1, it is imperative that ${{p}_{00}}+{{p}_{01}}=1$ and ${{p}_{10}}+{{p}_{11}}=1$.

Warner \cite{ref28} initially introduced the randomized response mechanism in 1965. Warner's model aligns with the particular instance in which ${{p}_{00}}={{p}_{11}}={{p}_{w}}$. In Warner's randomized response model, respondents provide their responses based on a random protocol, rather than answering questions directly. This random protocol may involve simple actions such as flipping a coin, rolling a die, or the outcomes of other random events. Consequently, respondents determine whether to provide truthful responses based on the randomly generated outcome. For instance, if the random result indicates ``yes'', respondents may be instructed to answer truthfully; however, if the outcome is ``no'', participants might be directed to respond in a predetermined manner, such as choosing between ``yes'' or ``no'' without revealing the actual circumstances. Hence, the design matrix of Warner's model ${{P}_{w}}$ is expressed as:
\begin{equation}
	{{P}_{w}}=\left( 
	\begin{matrix}
		{{p}_{w}} & 1-{{p}_{w}}  \\
		1-{{p}_{w}} & {{p}_{w}}  \\
	\end{matrix} \right).  \label{5}
\end{equation}

\section{The proposed scheme}
We will provide a comprehensive description of our scheme in this section. Firstly, the face image dataset is transformed into the face attribute database. Secondly, we impose differential privacy perturbations on the facial attribute database using the randomized response mechanism. Finally, the image synthesis technique is employed to generate the released facial image dataset.

\subsection{Construction of face attribute database}
At present, it is widely accepted that differential privacy guarantees should be offered when publishing individual-level image data. Generally, differential privacy mechanisms center on the introduction of controlled noise or perturbations during database querying or analysis, making it impossible to accurately infer individual data. This is achieved by randomizing the original data before the publication of the database by introducing noise, perturbations, or sampling. The current methods generally consider a single image as a database and define image pixels, the singular vectors of images, or the latent code representations of images as database records. This does not meet the criteria for differential privacy and fails to ensure semantic privacy and utility.

To comply with the concept of differential privacy, we will establish a face attribute database that aims to ensure privacy protection while facilitating the effective utilization of diverse attributes within facial images. For convenience, we directly utilize the attribute labels from the facial image dataset. These extracted attribute labels provide a crucial foundation of information for tasks such as facial recognition, facial verification, and facial image synthesis. Attribute labels typically do not contain the facial images themselves but rather comprise a series of abstract features, such as gender, age, and ethnicity. Attribute labels contain relatively few sensitive details. When performing data processing and analysis, utilizing attribute labels enhances control and management effectiveness.

\subsection{Face attribute perturbation}
This process yields the perturbed attribute database according to the randomized response mechanism, as defined below:

\textbf{Definition 3.} For a given ${\varepsilon}_{w} \ge 0$, a randomized response mechanism is considered ${\varepsilon}_{w} $-differentially private if, for any pair ${{x}_{i}}\in \{0,1\}$ and ${{x}_{j}}\in \{0,1\}$, and for any element $y\in \{0,1\}$,
\begin{equation}
	\Pr \left[ \operatorname{Res}\left( {{x}_{i}} \right)=y \right]\le {{e}^{{\varepsilon}_{w} }}\Pr \left[ \operatorname{Res}\left( {{x}_{j}} \right)=y \right]. \label{6}
\end{equation}

For the randomized response mechanism given by Eq. (4) to satisfy ${\varepsilon}_{w} $-differential privacy, the following conditions must be hold:
\begin{equation}
	{{p}_{00}}\le {{e}^{{\varepsilon}_{w} }}{{p}_{10}}, {{p}_{11}}\le {{e}^{{\varepsilon}_{w} }}{{p}_{01}}, {{p}_{01}}\le {{e}^{{\varepsilon}_{w} }}{{p}_{11}}, {{p}_{10}}\le {{e}^{{\varepsilon}_{w} }}{{p}_{00}}.   \label{7}
\end{equation}
Especially, the Warner's randomized response mechanism followed by design matrix ${{P}_{w}}$ in Eq. (5) satisfies ${\varepsilon}_{w}$-differential privacy, if
\begin{equation}
	{{p}_{w}}\le {{e}^{{\varepsilon}_{w} }}\left( 1-{{p}_{w}} \right), 1-{{p}_{w}}\le {{e}^{{\varepsilon}_{w} }}{{p}_{w}}.     \label{8}
\end{equation}

In the subsequent result, we will illustrate how Warner's randomized response mechanism achieves differential privacy.

\textbf{Lemma 1.} The randomized response method introduced by Warner \cite{ref28} conforms to ${\varepsilon}_{w} $-differential privacy, where
\begin{equation}
	{\varepsilon}_{w} =\max \left\{ \ln \frac{1-{{p}_{w}}}{{{p}_{w}}},\ln \frac{{{p}_{w}}}{1-{{p}_{w}}} \right\}.      \label{9}
\end{equation}

Here, ${p}_{w}$ denotes the probability that the attribute label remains unchanged following the implementation of the randomized response mechanism. 
As the value of ${p}_{w}$ decreases, the influence on the randomness of the attribute label increases. It should be noted that ${{p}_{w}}$ need to be selected carefully since it controls the privacy budget ${\varepsilon}_{w} $ that is different from $\varepsilon $ in Definition 1. 

In the case where ${{p}_{w}}={1}/{2}$, all rows of the design matrix exhibit identical values. This gives ${\varepsilon}_{w} =0$, consequently leading to a lack of meaningful insights to be aquaired. The optimal mechanism is not well-defined as ${{p}_{w}}$ approaches ${1}/{2}$.

If ${{0}<{p}_{w}}<{1}/{2}$ (or ${{1}/{2}<{p}_{w}}<{1}$), we can rearrange all responses so that ${{{y}'}_{i}}=1-{{y}_{i}}$. This aligns with the process of interchanging columns within the design matrix, resulting in ${{{p}'}_{w}}=1-{{p}_{w}}$, thereby ${1}/{2}<{{{p}'}_{w}}<{1}$ (or ${0}<{{{p}'}_{w}}<{1}/{2}$). For example, both the case ${{p}_{w}}=0.9$ and ${{p}_{w}}=0.1$ obtain $\ln 9$-differential privacy. Based on this, we can find that the value of ${\varepsilon}_{w} $ decreases as ${{p}_{w}}$ increases in the case where $0<{{p}_{w}}<{1}/{2}$, and the value of ${\varepsilon}_{w} $ increases as ${{p}_{w}}$ increases in the case where ${1}/{2}<{{p}_{w}}<1$.

It is evident that the case where ${1}/{2}<{{p}_{w}}<1$ fully aligns with the principles of differential privacy theory. Specifically, the lower the privacy budget is set, the higher the accuracy loss will be, and the greater the randomness in attribute label changes will be.  When $0<{{p}_{w}}<{1}/{2}$, reducing the privacy budget leads to decreased loss of accuracy and randomness in attribute label changes. This is also in line with the principles of differential privacy theory, given the relationship between ${\varepsilon}_{w} $ and ${p}_{w}$.
We only discuss the case ${0<{p}_{w}}<{1}/{2}$ in this paper.

Furthermore, we provide the structure of the design matrix aimed at obtaining optimal utility while adhering to the specified ${\varepsilon}_{w}$-differential privacy requirements.

\textbf{Lemma 2.} To maximize ${{p}_{w}}$ while satisfying $\varepsilon $-differential privacy, the design matrix ${{P}_{w}}$ should exhibit the following structure:
\begin{equation}
	{{P}_{w}}=\left( \begin{matrix}
		\frac{{{e}^{{\varepsilon}_{w} }}}{{{e}^{{\varepsilon}_{w} }}+1} & \frac{1}{{{e}^{{\varepsilon}_{w} }}+1}  \\
		\frac{1}{{{e}^{{\varepsilon}_{w} }}+1} & \frac{{{e}^{{\varepsilon}_{w} }}}{{{e}^{{\varepsilon}_{w} }}+1}  \\
	\end{matrix} \right).       \label{10}
\end{equation}
Proof. Suppose that $\frac{{{p}_{w}}}{1-{{p}_{w}}}=q$, we have $1<q\le {{e}^{{\varepsilon}_{w} }}$ to attain ${\varepsilon}_{w} $-differential privacy. In this case, the design matrix will possess the general form:
\begin{equation}
	P=\left( \begin{matrix}
		\frac{q}{1+q} & \frac{1}{1+q}  \\
		\frac{1}{1+q} & \frac{q}{1+q}  \\
	\end{matrix} \right).   \label{11}
\end{equation}
Let $f(q)=\frac{q}{1+q}$ be defined, then the derivative of $f(q)$, denoted as ${f}'(q)$, is given by ${f}'(q)=\frac{1}{{{(1+q)}^{2}}}>0$. Thereby, the maximum value of $f$ will be achieved if and only if $q={{e}^{{\varepsilon}_{w} }}$.

\subsection{Synthesize of face image dataset}
After the implementation of the random response mechanism to perturb the face attribute database, we utilized image synthesis techniques to release images. In this manner, this approach presents a challenge for an adversary to ascertain the inherent attributes of the original image through image analysis. The pivotal procedure involves directing the image generation process through the manipulation of the perturbed binary facial features. Therefore, our network comprises three primary components: an attribute classifier identified as $C$, a discriminator designated as $D$, and a generator denoted as $G$. Within the generator, there exists both an encoder, labeled as ${{G}_{enc}}$, and a decoder, represented by ${{G}_{dec}}$.

Given a face image $X$ in the spatial domain ${R}^{H\times W\times 3}$ featuring $n$ binary attributes represented by $a=\left[ {{a}_{1}},\cdots, {{a}_{n}} \right]$ and perturbed attribute labels denoted as ${a}'$, the encoder ${{G}_{enc}}$ is first employed to transform the face image $X$ into its latent representation $z={{G}_{enc}}(X)$, and then a synthesized image ${X}'={{G}_{dec}}\left( z,{a}' \right)$ is obtained by decoding $z$ conditioned on ${a}'$. Thus the whole process of $G$, expressed as ${X}'={{G}_{dec}}\left( {{G}_{enc}}\left( X \right),{a}' \right)=G\left( X,{a}' \right)$, is essential for synthesizing a novel image ${X}'$. Following adversarial learning involving generator $G$, the discriminator $D$ is implemented to evaluate the perceptual accuracy and visual authenticity of the synthesized image. The attribute classifier $C$ applies an attribute classification constraint to ${X}'$, effectively predicting the desired attributes of the synthesized image with accuracy.

Our network is constructed through the integration of the adversarial loss, the constraint on attribute classification, and the reconstruction loss. The detail of each loss will be described in the following.

(1) Adversarial Loss: The introduction of adversarial learning, which involves the interaction between the discriminator and generator, aims to enhance the visual fidelity of the synthesized outcomes. The adversarial losses can be distinctly specified as:
\[\mathcal{L}_{adv}^{G}=-{{E}_{{X}'\sim {{p}_{data}},{a}'\sim {{p}_{attr}}}}[D({X}')],\]
\[\mathcal{L}_{adv}^{D}=-{{E}_{X\sim {{p}_{data}}}}[D(X)]+{{E}_{{X}'\sim {{p}_{data}},{a}'\sim {{p}_{attr}}}}[D({X}')].\]
Here, ${{p}_{attr}}$ and ${{p}_{data}}$ are distributions of attributes and images, respectively. The training process is stabilized via WGAN-GP \cite{ref29}.

(2) Attribute Classification Loss: As previously mentioned, we aim to ensure that the synthesized images accurately possess the new attribute ${a}'$. Following this, we express the formulation of the attribute classification constraint as:
\[\mathcal{L}_{attr}^{G}={{E}_{X\sim {{p}_{data}},{a}'\sim {{p}_{attr}}}}[-\log ({{D}_{attr}}({a}'\left| G(X,{a}') \right.))].\]
Simultaneously, the attribute classifier should be trained on input images featuring their attributes to achieve higher classification accuracy,
\[\mathcal{L}_{attr}^{C}={{E}_{X\sim {{p}_{data}}}}[-\log ({{D}_{attr}}(a\left| X \right.))].\]

(3) Reconstruction Loss: To preserve the details regardless of attributes, the decoder is required to reconstruct $X$ by decoding the latent representation in alignment with attributes $a$. The formulation of the reconstruction loss term, using the ${{L}_{1}}$ norm, is expressed as:
\[\mathcal{L}_{rec}^{G}={{E}_{X\sim {{p}_{data}}}}[||X-G(G(X,a),a)|{{|}_{1}}].\]

(4) General Objective: Through the integration of the adversarial loss, the reconstruction loss, and the constraint on attribute classification, we give the target term for the encoder and decoder as
\[\underset{G}{\mathop{\min }}\,{{\mathcal{L}}_{G}}=\mathcal{L}_{adv}^{G}+{{\lambda }_{attr,g}}\mathcal{L}_{attr}^{G}+{{\lambda }_{rec}}\mathcal{L}_{rec}^{G},\]
and the objective item for the attribute classifier and discriminator as
\[\underset{D,C}{\mathop{\min }}\,{{\mathcal{L}}_{D,C}}=\mathcal{L}_{adv}^{D}+{{\lambda }_{attr,c}}\mathcal{L}_{attr}^{C},\]
where the hyperparameters ${\lambda }_{attr,g}$, ${\lambda }_{rec}$, and ${\lambda }_{attr,c}$ are utilized to adjust the balance of these losses.

\section{Experimental results}
\subsection{Setup}
(1) Dataset: We utilize the CelebFaces Attributes (CelebA) as the dataset, which is a substantial collection of face attributes comprising more than 200,000 images, each labeled with 40 binary attributes. The CelebA dataset encompasses both original images and cropped, aligned facial images. Formally, we employ a dataset comprising 182,000 images categorized into training, validation, and testing sets: 182,000 images for training, 637 images for validation, and the remainder for assessment during testing. Taking into account diverse attributes across the face, we select two local attributes: ``Bangs'' and ``Blond Hair'', and three global attributes: ``Male'', ``Pale Skin'', and ``Young''. The original facial images are resized to $128\times 128$ pixels.

\begin{figure}[htbp]
	\centering
	\includegraphics[width =\linewidth]{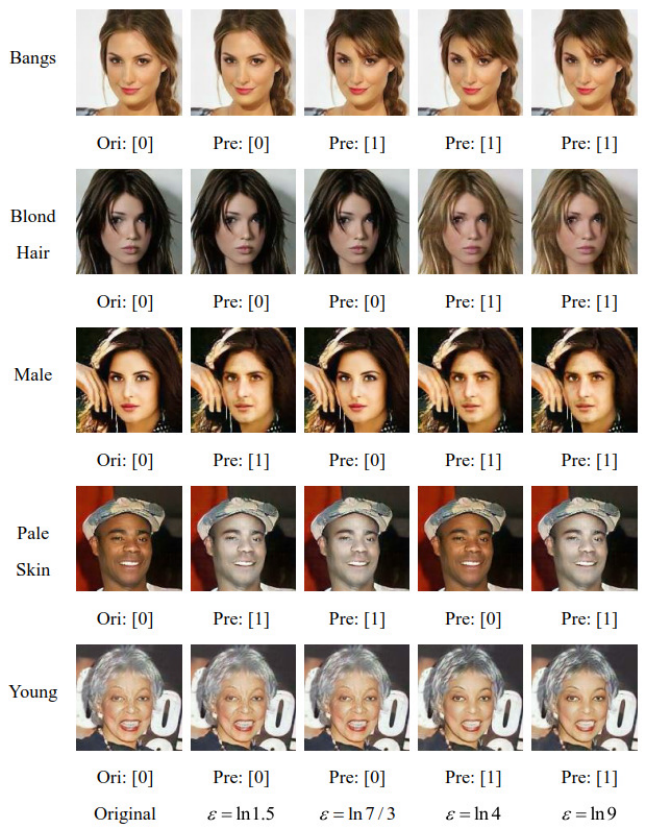}
	\caption{Visual results for a single attribute. Original faces are shown in the first column, while the rest columns display corresponding faces with a single perturbed attribute under different $\varepsilon$. From top to bottom, the attributes are ``Bangs'', ``Blond Hair'', ``Male'', ``Pale Skin'', and ``Young'', respectively.}
	\label{Fig.2(1)}
\end{figure}

\begin{figure}[htbp]
	\includegraphics[width =\linewidth]{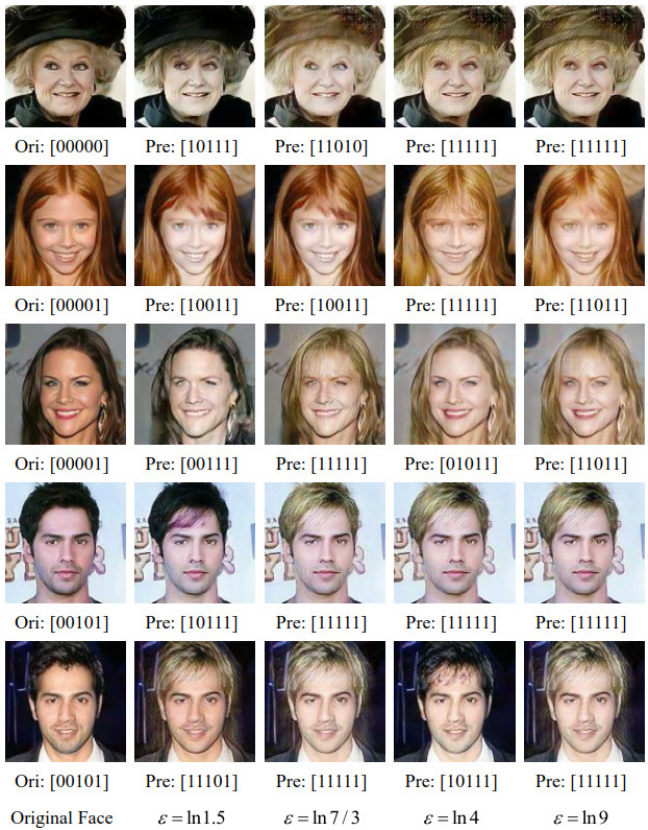}
	\caption{Visual results for multiple attributes. Original faces are shown in the first column, while the rest columns display corresponding faces with multiple perturbed attributes under different $\varepsilon$.}
	\label{Fig.2}
\end{figure}

(2) Implementation Details: To gain a comprehensive understanding of the differential privacy mechanism employed in our scheme, we devise experiments to examine and analyze the influence of the privacy budget parameter ${\varepsilon}_{w}$ on face attributes. Initially, we extract the face attribute database from the image dataset and then obtain the perturbed attribute database using the randomized response mechanism. Then we increase ${{p}_{w}}$ from 0.6 to 0.9 in increments of 0.1, and adjust each attribute for each ${{p}_{w}}$. Accordingly, we increase ${\varepsilon}_{w}$ from $\ln 1.5$ to $\ln 9$, based on Eq. (9). The model utilized for image synthesis is trained using the Adam optimizer with ${{\beta}_{1}}=0.5$ and ${{\beta}_{2}}=0.99$. The optimal settings of hyperparameters associated with the attribute classification loss and the identity loss are ${{\lambda }_{attr,g}}=10$, ${{\lambda }_{attr,c}}=1$, and ${{\lambda }_{rec}}=100$. The initial learning rate is set to 0.0002, with a batch size of 32. To evaluate the privacy and utility accuracy for face attributes, we choose an attribute classifier, i.e., ResNet \cite{ref30}, to estimate the intrinsic attributes of the generated face images. The attribute classifier is trained on the CelebA dataset, achieving an average accuracy of 90.9$\%$ for each attribute on the CelebA testing set.

\begin{figure}[htbp]
	\includegraphics[width =\linewidth]{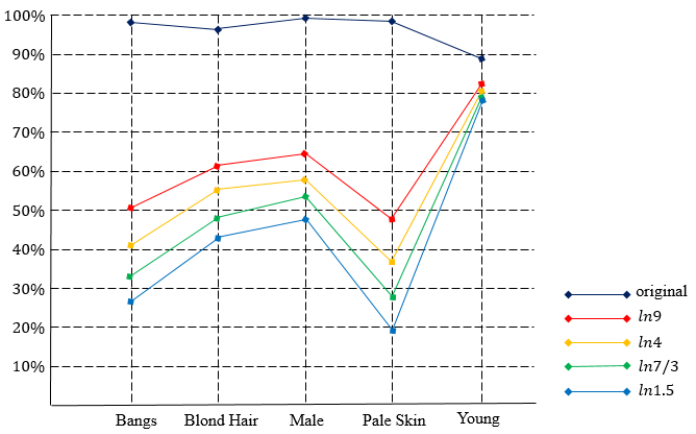}
	\caption{Attribute privacy protection performance. The accuracies for every attribute are evaluated on both the original images and the synthesized results under varying ${\varepsilon}_{w} $ values.}
	\label{Fig.3}
\end{figure}

\begin{figure}[htbp]
	\includegraphics[width =\linewidth]{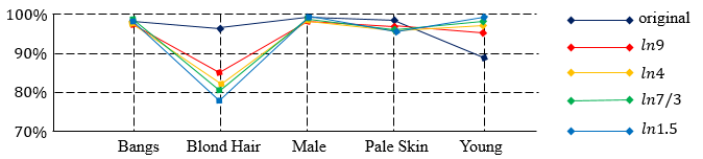}
	\caption{Attribute utility performance. The accuracies for each attribute are assessed on both the original images and the synthesized outcomes with different ${\varepsilon}_{w} $ values.}
	\label{Fig.4}
\end{figure}

\subsection{Evaluation on privacy and utility for attribute}
(1) Evaluation on the performance of attribute privacy: To demonstrate the protective efficacy of our scheme for a specific attribute, we show samples with different privacy budgets in Fig. 2. ``Ori'' signifies the accuracy metrics of the employed classifier on the input data. ``Pre'' represents the classification results on the synthesized images. The values 0 and 1 indicate the possible values of the sensitive attribute. Based on the results illustrated in Fig. 2, we can conclusively assert that our scheme can prevent a strong adversary, who knows all but one attribute in the database, from further inferring the last attribute.

To demonstrate the protective capacity of the proposed scheme for multiple attributes, we present the results of five attributes: ``Bangs'', ``Blond Hair'', ``Male'', ``Pale Skin'', and ``Young'' with different privacy budgets in Fig. 3. As we observe, the proposed scheme exhibits a notably robust capacity to safeguard multiple attributes. From a visual perspective, there is no apparent decrease in the quality of the face images. Meanwhile, it presents a challenge for us to discern which original attributes are concealed. This reduces the threat that an attacker poses to an individual's privacy by accessing a sufficient number of attributes.
 
\begin{figure*}[htbp]
	\centering
	\vspace{1.0em}
	\includegraphics[width=0.9\linewidth]{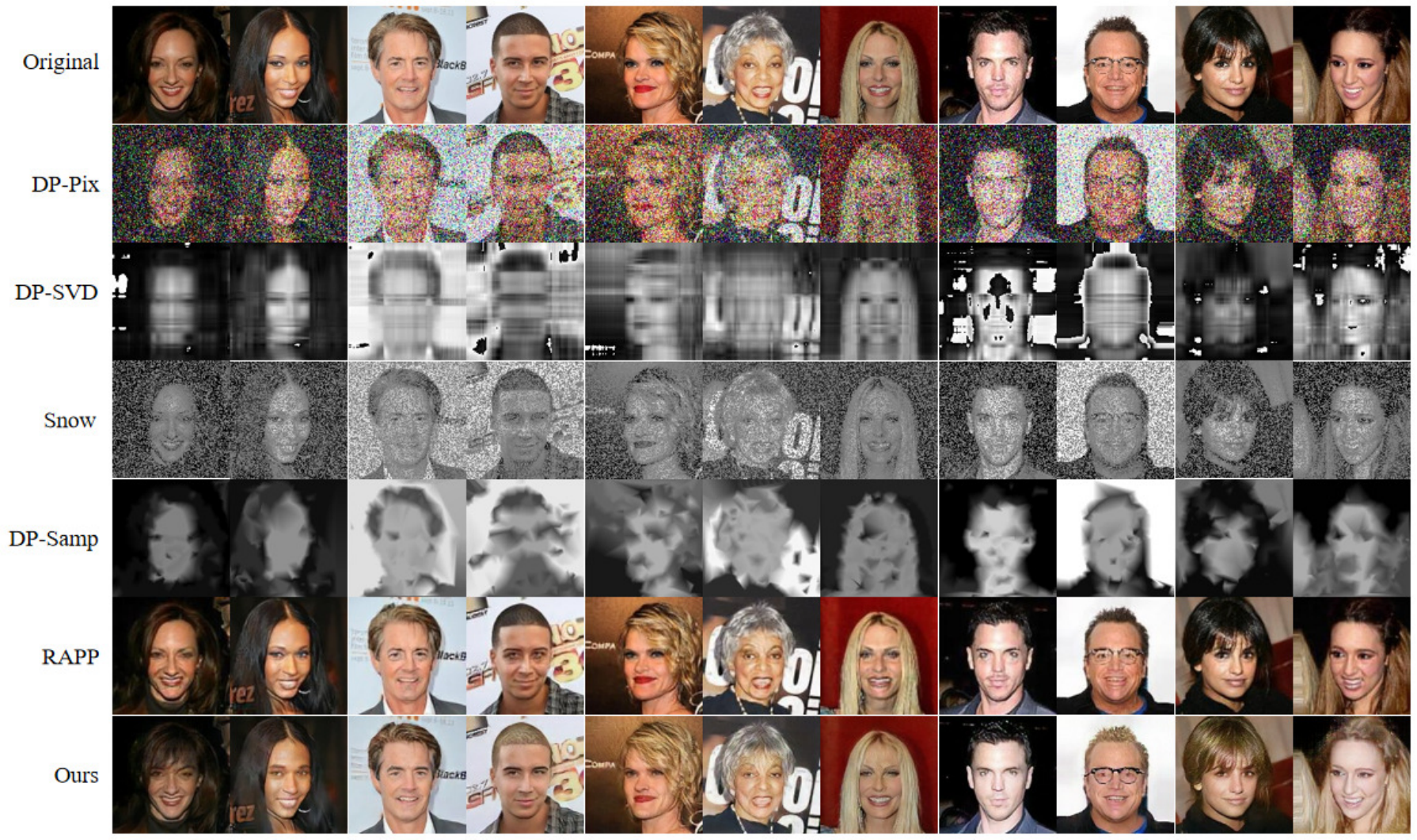}
	\caption{Qualitative assessment in comparison to other approaches. Original faces, faces generated by DP-Pix, DP-SVD, Snow, DP-Samp, RAPP, and our scheme (${\varepsilon}_{w} =ln1.5$) are arranged from the topmost to the bottom.}
	\label{Fig.5}
\end{figure*}

\begin{table*}[htbp]
	\centering
	\caption{Quantitative assessment of CelebA datasets using PSNR, SSIM and LPIPS.} 
	\label{Table I} 
	\setlength{\tabcolsep}{22pt}
	\renewcommand{\arraystretch}{1.3}
	\begin{tabular}{{llllll}} 
		\hline 
		{} & DP-Pix  & DP-SVD  & Snow  & DP-Samp   & Ours \\ 
		\hline 
		PSNR$\uparrow $                  & 12.3949 & 15.0991  & 13.9066  & 17.5281    & \textbf{18.6680}   \\ 
		
		SSIM$\uparrow $                  & 0.1452 & 0.4930  & 0.3043  & 0.4996    & \textbf{0.7427}   \\ 
		
		LPIPS$\downarrow $                  & 0.7784 & 0.5767  & 0.6449  & 0.5244    & \textbf{0.2299}   \\ 
		\hline 
	\end{tabular}
\end{table*}

In assessing the effectiveness of protecting face attributes, we regard the attribute classifier ResNet as an attacker to ascertain the attributes present in the synthesized faces. We employ accuracy (Acc) as the primary classification metric for evaluating the protection of attribute privacy. When the Acc value of the classifier for the protected image is low, we consider privacy protection successful; otherwise, it is not successful. According to the results depicted in Fig. 4, it is apparent that our scheme substantially reduces the performance of the attribute classifier across various ${\varepsilon}_{w}$, thus illustrating its efficacy in safeguarding attribute privacy. On the one hand, the average accuracy of the classifier on the test data decreases substantially. On the other hand, the accuracy values of the classifier for different attributes increase as ${\varepsilon}_{w}$ increases from $\ln 1.5$ to $\ln 9$. In general, an increase in the privacy budget can improve the accuracy because a higher privacy budget has less impact on the randomness of the attribute database. Among these selected attributes, ``Bangs'' and ``Pale Skin'' exhibit the highest concealment, as evidenced by their classification accuracy, while the attribute ``Young'' displays minimal degradation in classification performance.

(2) Evaluation on the performance of attribute utility: We also use accuracy (Acc) as our primary classification metric for the evaluation of the utility of the attribute dataset. When the Acc value of the classifier for the protected image is high, we consider that the data utility is large; and vice versa. Fig. 5 plots the curves for Acc under different $\varepsilon$, indicating that the visual utility does not change drastically with the increase in the privacy budget. Among these selected attributes, the utility for all the attributes remains at a high level, while the attribute ``Blond Hairs'' has a slight decrease as the privacy budget decreases, achieving an average accuracy of around 81.4$\%$.

\subsection{Comparisons with SOTA schemes}
In this section, we conduct an empirical analysis to present a comparative evaluation. In order to assess the effectiveness of the proposed scheme, we perform a comparative analysis with five other schemes: (1) DP-Pix \cite{ref13}: DP-Pix achieves differential privacy by directly perturbing super-pixels. The method employs pixelization to address the high sensitivity in image publication, and integrates the concept of neighboring databases into the image domain via the definition of $m$-neighborhood. The parameters are assigned default values in accordance with recommended experimental settings, privacy parameter $\varepsilon =0.5$, the number of allowed different pixels $m=16$, grid cell length $b=16$. (2) DP-SVD \cite{ref15}: DP-SVD mechanism perturbs perceptual features extracted from the input image. This method first conducts singular value decomposition (SVD) on an input image and subsequently attains metric privacy through the perturbation of the $k$ highest singular values. The parameters are configured with default values, i.e., $\varepsilon =0.5$ and $k=4$. (3) Snow \cite{ref14}: Snow utilizes pixel-level noise by randomly reallocating pixel intensities to a fixed value 127 for grayscale images. This approach attains $(0,\delta )$-DP, where $\delta =1-p$, and $p$ determines the proportion of modified pixels. A value of $p$ is set to be 0.1 according to the recommended experiment setting. (4) DP-Samp \cite{ref31}: DP-Samp is an adaption of the video sanitization method technique \cite{ref32} designed to protect up to $m$ pixels within individual images, consisting of pixel clustering, budget allocation, pixel sampling with top $k$ pixel values, and interpolation. We set parameters $k=5$, $m=5$, and $\varepsilon =0.1$. (5) RAPP \cite{ref40}: RAPP comprises two distinct modules: an attribute obfuscator, which employs a password mechanism to generate unpredictable labels that conceal specific attributes, and a generator derived from an attribute adversarial network.

Fig. 6 shows the results of six methods on face images from the CelebA dataset. As can be seen, all methods provide different levels of privacy protection for images. However, RAPP fails to offer demonstrable privacy guarantees, even though it slightly changes attributes and yields significant utility. DP-Pix employs standard differential privacy methods that require more perturbation, consequently resulting in lower utility. By introducing singular value decomposition, DP-SVD only yields the largest singular values to guarantee the performance of the mechanism, which leads to a greater loss of image structural information, ultimately yielding low utility. Snow employs a mechanism derived from salt-and-pepper noise, commonly referred to as ``snow'', to introduce pixel-level noise, thereby increasing the risk of reidentification. DP-Samp utilizes pixel subsampling techniques, causing increased information loss and subsequently a reduction in image quality. Comparatively, our method exhibits notably superior utility for all images, simultaneously ensuring provable privacy protection.

Besides subjective visual comparisons, we also use image quality assessment metrics to give a quantitative analysis. To measure image quality, we adopt PSNR, SSIM, and LPIPS (learned perceptual image patch similarity distance) \cite{ref33}. PSNR and SSIM are the predominant metrics employed for evaluating the semantic similarity assessment between the reference image and the reconstructed image. LPIPS is utilized for assessing visual similarity, demonstrating a closer alignment with human perceptual judgments compared to conventional metrics. Typically, the higher values of PSNR/SSIM and the lower values of LPIPS signify superior image quality. The average PSNR, SSIM, and LPIPS values of the results under the CelebA dataset are listed in Table 1. Here, RAPP is not a DP-based method that makes minor alterations to facial attributes. We exclusively employ these metrics to assess the efficacy of our scheme in comparison to existing DP-based methods concerning image utility. It can be observed that the proposed scheme exhibits a considerable advantage in visual similarity and maintains the structural coherence of the original image to a considerable degree. Table 1 presents the comprehensive quantitative results, indicating that current DP-based methods do not succeed in enhancing the tradeoffs between utility and privacy protection.

\section{Conlusion}
In this paper, our primary emphasis is directed towards the limitations of DP-based face image privacy protection. We investigate strategies for providing differential privacy for whole face image datasets. Our scheme consists of three stages: construction of the face attribute dataset, face attribute perturbation, and synthesis of the face image dataset. In our scheme, differential privacy perturbation is directly implemented on the face attribute dataset to guarantee its accessibility while preserving the attributes. Experiments demonstrate the advantage and reliability of our scheme concerning both privacy preservation and image utility. Moreover, the results indicate satisfactory performance when compared to SOTA schemes. In subsequent research, we intend to expand the scope of this study to video datasets. For certain (continuous) frames containing sensitive facial images, it is interesting to achieve temporally consistent protection measures.

\section*{Acknowledgments}
This work was supported by the Excellent Postdoctoral Program of Jiangsu Province (No. 2023ZB435 ).

\newpage
\bibliographystyle{IEEEtran}
\bibliography{IEEEexample}
%



\end{document}